# Learning Measurement Models for Unobserved Variables

Ricardo Silva, Richard Scheines, Clark Glymour and Peter Spirtes*
Center for Automated Learning and Discovery
Carnegie Mellon University
Pittsburgh, PA 15213

## Abstract

Observed associations in a database may be due in whole or part to variations in unrecorded ("latent") variables. Identifying such variables and their causal relationships with one another is a principal goal in many scientific and practical domains. Previous work shows that, given a partition of observed variables such that members of a class share only a single latent common cause, standard search algorithms for causal Bayes nets can infer structural relations between latent variables. We introduce an algorithm for discovering such partitions when they exist. Uniquely among available procedures, the algorithm is (asymptotically) correct under standard assumptions in causal Bayes net search algorithms, requires no prior knowledge of the number of latent variables, and does not depend on the mathematical form of the relationships among the latent variables. We evaluate the algorithm on a variety of simulated data sets.

## 1 Introduction

A great deal of contemporary science has two striking features. First, its goals and results are typically about causation or composition—what minerals compose a soil sample; what mechanism regulates expression of a particular gene; what effect does low level lead exposure have on children's intelligence? Second, scientific data, the measurements and observations upon which hypotheses are discovered, tested, and refined, are indirect. They are not measurements of the scientifically important causal features themselves, but only of their more easily observed effects. We do not measure the mineral composition of a soil sample directly, we measure its spectra; we do not measure gene regulation directly, we measure light intensities on microarray chips; we do not measure children's exposure to lead, we measure the concentration of lead in their baby teeth; and so on. These two aspects of modern science pose a fundamental problem for computer aided data analysis. Our evidence is a sample of values for a set of observed variables; what we want to infer involves causation or composition among "latent" variables, i.e., variables whose values are not recorded. Inevitably, assumptions must be made and models built that connect evidence to theory, but finding the right assumptions for the scientific task is not obvious. Sometimes the assumptions are too weak and radically underdetermine the underlying structure. Principal components methods are an example. Sometimes the assumptions are arbitrary, as with the choice of particular rotations in factor analysis (Bartholomew, et al., 2002). Sometimes they are too strong for most scientific contexts, as with independent components analysis, in which the underlying signal sources are assumed to be independent in probability, and therefore also causally independent (Hyvarinen, et al., 2001). For a variety of reasons it has proved difficult to exploit Bayes nets in searching for the causal or compositional structure among a set of latent variables: the likelihood surface of "latent variable models" is very irregular (Geiger, et al., 2001); the models do not always have a well-defined dimension, the space of models is infinite, etc. One important fact is known however: if for each latent variable there are at least three measured effects, and these measures are otherwise suitably unconfounded – or "pure" in a sense we make precise below – then standard Bayes net search procedures can be correctly applied to obtain information about the connections among the latent variables (Spirtes, et al., 2000; ch. 12). If, therefore,

*The last three authors are also affiliated with the Department of Philosophy, Carnegie Mellon University. Research for this paper was supported by grants NCC2-1377, NCC2-1295 and NCC1-1227 to IHMC at the University of West Florida.

there were a correct algorithm for locating such sets of measured variables when they exist, subject only to the Markov and Faithfulness assumptions (Spirtes, et al., 2000) and perhaps the assumption of particular distribution families (e.g., Gaussian, multinomial, etc.), a principled method for discovering latent structure would be available for a class of problems. We describe such an algorithm for cases in which observed variables depend linearly on latent variables, assuming nothing about the nature of the relationships among the latents[1].

## 2 The Set-Up

Our procedure first finds disjoint subsets of measured variables such that members of each subset have a single latent common cause, but may be otherwise confounded or impure. Each subset is refined to eliminate confounded variables, and the procedure returns an equivalence class of measurement models, a pure *measurement pattern*. No a priori choice of the number of latent factors is made. Provided the assumptions of the algorithm are satisfied and all statistical decisions are made correctly, it provably finds correctly specified purified measurement models. We describe the essentials of the algorithm here. Proofs are given in (Silva, et al., 2003).

### 2.1 Definitions

**Definition 1 (Measurement model)** *A directed acyclic graph (DAG) containing a set of latent variables* $\mathbf{L}$, *a set of error variables* $\epsilon$, *a set of observed variables* $\mathbf{O}$, *two sets of edges* $\mathbf{E_O}$ *and* $\mathbf{E}_\epsilon$, *forms a measurement model* $M(\mathbf{L}, \mathbf{O}, \epsilon, \mathbf{E_O}, \mathbf{E}_\epsilon)$ *if each latent in* $\mathbf{L}$ *is a parent of at least one variable in* $\mathbf{O}$, *none of the observed variables is a parent of any variable in* $\mathbf{L} \cup \epsilon$, *all nodes in* $\mathbf{O}$ *are children of some node in* $\mathbf{L}$, *any node in* $\epsilon$ *is a common parent of at least two nodes in* $\mathbf{O}$ *and is d-separated from every element of* $\mathbf{L}$ *given the empty set. All edges in* $\mathbf{E_O}$ *are directed into* $\mathbf{O}$ *and all edges in* $\mathbf{E}_\epsilon$ *are directed from* $\epsilon$ *into* $\mathbf{O}$.

The definition of a measurement model specifies in which way observed variables are indicators of latent factors but does not consider how such factors are related. Nodes in $\epsilon$ represent dependent latent errors, analogous to error variables in regression. Instead of explicitly showing such error nodes in our figures, we link any pair of observed nodes that share a common error parent with a double-headed edge. We restrict our discussion to measurement models in which each observed variable is a linear function of its parents plus additive noise: $O_i = \sum_j \lambda_{ij} P_{ij} + e_i$, where each $P_{ij}$ represents a parent of the observed $O_i$, and $e_i$ is independent of all variables in the model other than $O_i$.

**Definition 2 (Pure measurement model)** *A measurement model* $M(\mathbf{L}, \mathbf{O}, \epsilon, \mathbf{E_O}, \mathbf{E}_\epsilon)$ *is pure if and only if for every* $O_i \in \mathbf{O}$, $O_i$ *has a single latent parent* $L_i$ *and* $L_i$ *d-separates* $O_i$ *from every element in* $(\mathbf{L} - L_i) \cup (\mathbf{O} - O_i)$.

Notice that, in a pure measurement model, $\epsilon = \emptyset$ and $\mathbf{E}_\epsilon = \emptyset$.

**Definition 3 (Latent variable graph)** *Given a set of latent variables* $\mathbf{L}$, *a set of error variables* $\epsilon$, *a set of observed variables* $\mathbf{O}$, *three sets of edges* $\mathbf{E_O}$, $\mathbf{E_L}$ *and* $\mathbf{E}_\epsilon$, *a latent variable graph* $G(\mathbf{L}, \mathbf{O}, \epsilon, \mathbf{E_L}, \mathbf{E_O}, \mathbf{E}_\epsilon)$ *is a directed acyclic graph, all edges in* $\mathbf{E_L}$ *have both endpoints in* $\mathbf{L}$, *and the directed acyclic graph defined by the tuple* $(\mathbf{L}, \mathbf{O}, \epsilon, \mathbf{E_O}, \mathbf{E}_\epsilon)$ *forms a measurement model.*

Given a latent variable graph, the tuple $(\mathbf{L}, \mathbf{O}, \epsilon, \mathbf{E_O}, \mathbf{E}_\epsilon)$ is its measurement model. We will say a latent variable graph is pure if its measurement model is pure, and that a linear latent variable graph is a latent variable graph with a linear measurement model. A purification of a latent variable graph is a pure latent variable graph obtained by possibly deleting some of the observed variables.

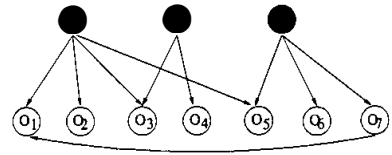

Figure 1: The graph in this figure is not a pure model: $O_1$ and $O_7$ are d-connected given their latent parents, $O_3$ and $O_5$ have more than one parent.

The graph in Figure 1 has a purification containing variables $\{O_2, O_4, O_6, O_7\}$ and any subset of this set.

### 2.2 Assumptions

**Definition 4 (Purifiable linear latent variable graph)** *A purifiable linear latent variable graph* $G(\mathbf{L}, \mathbf{O}, \epsilon, \mathbf{E_L}, \mathbf{E_O}, \mathbf{E}_\epsilon, \mathbf{G_S})$ *is a graphical model such that the tuple* $(\mathbf{L}, \mathbf{O}, \epsilon, \mathbf{E_L}, \mathbf{E_O}, \mathbf{E}_\epsilon)$ *is a linear latent variable graph and* $\mathbf{G_S}$ *is a non-empty set of purifications of* $G$ *such that, in every graph* $G_S \in \mathbf{G_S}$, *all latent nodes have at least three observed children in* $G_S$.

---
[1]Althought we are not assuming linearity among latents for the measurement model discovery problem, we do not know any consistent algorithm for finding causal models among continuous latents without assuming linearity.

The motivation for requiring at least three observed children per latent in purifications of $G$ arises from constraints on identifiability. This will be evident in the next sections, where we introduce an algorithm for learning families of measurement models (*equivalence classes*) that fit a given covariance matrix $\Sigma$ of a set of variables $\mathbf{O}$. The assumptions under which the algorithm is correct are:

- the observed variables $\mathbf{O}$ are continuous;

- $\Sigma$ is faithfully generated by an unknown purifiable linear latent variable graph $G(\mathbf{L}, \mathbf{O}, \epsilon, \mathbf{E_L}, \mathbf{E_O}, \mathbf{E_\epsilon}, \mathbf{G_S})$;

- the distributions of $\mathbf{O}$, $\mathbf{L}$ and $\epsilon$ have second moments;

We assume that the measurement model is linear, but we do not assume that the relations between the latents are linear, nor do we assume anything about the family of probability distributions over $\mathbf{O}$, $\mathbf{L}$ or $\epsilon$.

## 3 Equivalence classes

Search algorithms should recognize in their output alternative models that cannot be distinguished given the assumptions and the marginal probability distribution on the observed variables. For instance, *patterns* (Pearl, 2000) represent d-separation equivalence over DAGs. Analgously, our procedure ought to output equivalence classes of indistinguishable measurement models. Accordingly, the output of the main algorithm introduced in the next section is a *measurement pattern* $MM_G$, a graphical object with directed and undirected edges that represents an equivalence class of measurement models. $MM_G$ has the following properties:

- the graph $MM_G$ has a set $\mathbf{T}$ of latent variables and observed variables $\mathbf{O'} \subseteq \mathbf{O}$, where $\mathbf{O}$ is the original set given as input. Notice that we denote latents in the pattern by $\mathbf{T}$ instead of $\mathbf{L}$, because obtaining a one-to-one mapping from one set to the other is not guaranteed;

- every latent has at least two children;

- some pairs of observed variables may be connected by an undirected edge. Some pairs of latents are connected by an undirected edge. No latents have parents;

- there are no error nodes;

Let $G(\mathbf{L}, \mathbf{O}, \epsilon, \mathbf{E_L}, \mathbf{E_O}, \mathbf{E_\epsilon}, \mathbf{G_S})$ be a purifiable linear latent variable graph. Then $MM_G$ represents possible measurement models such that the measurement model of every $G_S \in \mathbf{G_S}$ is a subgraph of $MM_G$.

Our search problem can be seen as an unusual *clustering* problem. Clusters can overlap in general measurement models. Clusters cannot overlap in pure measurement models. Sometimes we will refer to the elements of $\mathbf{G_S}$ as *solution graphs*, because they can be identified as demonstrated later, while this is not usually the case for $G$. Unrepresented measurement error is implicit in the parameterization of the model.

## 4 An algorithm for learning measurement patterns and models

The algorithm here described builds a measurement pattern of a unknown purifiable linear latent variable graph with a known observed covariance matrix $\Sigma$ by evaluating the validity of *tetrad constraints* among sets of four variables. Given the covariance matrix of four random variables $\{A, B, C, D\}$, we have that zero, one or three of the following constraints may hold:

$$\begin{aligned}
\sigma_{AB}\sigma_{CD} &= \sigma_{AC}\sigma_{BD} \\
\sigma_{AC}\sigma_{BD} &= \sigma_{AD}\sigma_{BC} \\
\sigma_{AB}\sigma_{CD} &= \sigma_{AD}\sigma_{BC}
\end{aligned}$$

Statistical tests for tetrad constraints or vanishing tetrad differences are straightforward assuming normal covariates. The constraints can be tested for a larger family of distributions using fourth moments (Bollen, 1990). Their value lies in the fact that various simple Bayes net structures imply characteristic subsets of possible tetrad constraints for systems in which observed variables depend linearly on latents: a single latent cause of four observed variables implies all three vanishing tetrads; a single latent cause of three observed variables and another latent cause of a fourth observed variable, implies all three vanishing tetrads, no matter how the latents are related; a single latent cause of two observed variables and another latent cause of two other observed variables, implies exactly one vanishing tetrad, etc. (Glymour, et al., 1987).

### 4.1 Clustering and impurity identification

The function $TetradScore(\mathbf{Set}; \Sigma)$ counts the number of tetrad constraints that hold among elements in $\mathbf{Set}$, which have a covariance matrix as a submatrix of $\Sigma$, and where for no triple $\{X, Y, Z\} \subset \mathbf{Set}$ does $\rho_{XY.Z} = 0$ (the partial correlation of $X$ and $Y$ given $Z$ vanishes). If for some triplet we have $\rho_{XY.Z} = 0$, the $TetradScore$ is defined to be zero. Given the covariance matrix of a set of variables as an input, in outline the procedure is:

1. identify which variables are uncorrelated; such variables cannot be in the same cluster;

2. identify which pairs of variables $(X, Y)$ cannot form a one-factor model with some other pair. If it is not possible to find such a one-factor model, $X$ and $Y$ cannot be part of any graph in $\mathbf{G_S}$ at the same time, or otherwise we would be able to construct such a one-factor model (for instance, with two other elements from the cluster of $X$, if $X$ and $Y$ are not in the same cluster);

3. decide which pairs of variables $\{X, Y\}$ should not be in the same cluster by evaluating the predicate $Unclustered(\{X, A, B\}, \{Y, C, D\}; \Sigma)$, as defined in Table 1. Here, variables $\{A, B, C, D\}$ are other variables in the covariance matrix;

4. identify cliques formed by variables where no pair was labeled as incompatible by any of the three criteria above.

Table 1: Returns true only if no variable in $\mathbf{O_1}$ has a common parent with any variable in $\mathbf{O_2}$. The symbol $\rho_{O_x O_y . O_z}$ represents the partial correlation of $O_x$ and $O_y$ conditioned on $O_z$.

$Unclustered(\mathbf{O_1} = \{O_a, O_b, O_c\}, \mathbf{O_2} = \{O_x, O_y, O_z\}, \Sigma))$
if $\forall \{O_1, O_2\} \in \mathbf{O_1} \times \mathbf{O_2}, O_1$ is uncorrelated with $O_2$
**return** true
**else return**
$\forall O_x, O_y \in \mathbf{O_1} \cup \mathbf{O_2}, \rho_{O_x O_y} \neq 0$ and
$\forall O_x, O_y, O_z \in \mathbf{O_1} \cup \mathbf{O_2}, \rho_{O_x O_y . O_z} \neq 0$ and
$\forall V \in \mathbf{O_1}, \sigma_{V O_x} \sigma_{O_y O_z} = \sigma_{V O_y} \sigma_{O_x O_z} = \sigma_{V O_z} \sigma_{O_x O_y}$ and
$\forall V \in \mathbf{O_2}, \sigma_{V O_a} \sigma_{O_b O_c} = \sigma_{V O_b} \sigma_{O_a O_c} = \sigma_{V O_c} \sigma_{O_a O_b}$ and
$\forall \{O_i, O_j\} \subset \mathbf{O_1}, \{O_p, O_q\} \subset \mathbf{O_2},$
$\sigma_{O_i O_p} \sigma_{O_j O_q} = \sigma_{O_i O_q} \sigma_{O_j O_p} \neq \sigma_{O_i O_j} \sigma_{O_p O_q}$

When all relationships are linear, there is a fairly intuitive explanation for the $Unclustered$ test: if all three tetrads hold among elements in $\{V_1, V_2, V_3, V_4\}$, then there is some common ancestor d-separating such elements (assume for purpose of illustration that such common cause is not in $\{V_1, V_2, V_3, V_4\}$). Assume all three tetrads hold in $\{X, A, B, Y\}$ and $\{X, Y, C, D\}$, but $\sigma_{XY} \sigma_{AC} \neq \sigma_{XA} \sigma_{YC}$. If $X$ and $Y$ had a common parent, then this common parent would have to d-separate every member of $\{X, Y, A, B, C, D\}$, and therefore all tetrad constraints would hold in this set, which means $\sigma_{XY} \sigma_{AC} = \sigma_{XA} \sigma_{YC}$. Contradiction. A full proof for the non-linear latent structure case is given in (Silva et al., 2003). We illustrate the essential features of the procedure above outlined, which we will call FindMeasurementPattern, in Figure 2.

Figure 2a shows the true graph that is unknown to the algorithm. Initially, in (2b) we create a complete graph where all observed variables are vertices. In 2c, all edges in $\{1, 2, 3, 4\} \times \{9, 10, 11\}$ are removed because such sets are uncorrelated. In Figure 2d, other edges are removed because of the $Unclustered$ test. For example, $Unclustered(\{1, 2, 3\}, \{6, 7, 8\}; \Sigma)$ will hold. Since the pair $\{3, 5\}$ could not satisfy the second criterion of the sketch given above, we represent this failure by a dotted edge in Figure 2d. Next, we first separate the graph into components consisting of solid edges only, as in Figure 2e. All maximal cliques are generated for each of these components, generating three clusters in our example. Another graph is generated using these clusters (Figure 2f), and the dotted edge from the previous step is added back forming the undirected edge between $\{3, 5\}$. It remains to decide which latents in this graph should be linked. For each pair of latents, this is done by finding three indicators $\{O_1, O_2, O_3\}$ of the first latent, three indicators $\{O_4, O_5, O_6\}$ from second, and adding the edge between latents if $Unclustered(\{O_1, O_2, O_3\}, \{O_4, O_5, O_6\}; \Sigma)$ holds. If there is no such pair, then these latent will not be linked. In our example, all latents are linked. The resulting theorem follows with probability 1 (Silva et al., 2003):

**Theorem 1** *Let $G(\mathbf{L}, \mathbf{O}, \epsilon, \mathbf{E_L}, \mathbf{E_O}, \mathbf{E_\epsilon}, \mathbf{G_S})$ be the purifiable linear latent variable graph that generates the covariance matrix $\Sigma$ of a set of observed random variables $\mathbf{O}$. Then, $G$ will be in the measurement equivalence class $MM(\mathbf{O}, \Sigma)$, and such class will be given by the measurement pattern obtained throught FindMeasurementPattern$(\mathbf{O}, \Sigma)$.*

### 4.2 Purification

A measurement pattern is not a measurement model, but it is possible to find all pure measurement models of the unknown true graph from the measurement pattern. Let $M_G^H(T, L) = true$ if and only if all children of node $T$ in graph $G$ are children of node $L$ in graph $H$ (i.e., such children have the same name). We define the relationship $=_{MM}$ for two latent variable graphs $G_1(\mathbf{L_1}, \mathbf{O_1}, \epsilon_1, \mathbf{E_{L_1}}, \mathbf{E_{O_1}}, \mathbf{E_{\epsilon_1}})$ and $G_2(\mathbf{L_2}, \mathbf{O_2}, \epsilon_2, \mathbf{E_{L_2}}, \mathbf{E_{O_2}}, \mathbf{E_{\epsilon_2}})$ as $G_1 =_{MM} G_2$ if and only if $\mathbf{O_1} = \mathbf{O_2}$ and for each $L_1 \in \mathbf{L_1}$ there exists an unique $L_2 \in \mathbf{L_2}$ such that $M_G^{G_2}(L_1, L_2) = true$ and $M_{G_2}^{G_1}(L_2, L_1) = true$. For two sets of latent variable graphs $\mathbf{G_1}$ and $\mathbf{G_2}$, we have $\mathbf{G_1} =_{MM} \mathbf{G_2}$ if for every $G_1 \in \mathbf{G_1}$ there is an unique $G_2 \in \mathbf{G_2}$ such that $G_1 =_{MM} G_2$ and $|\mathbf{G_1}| = |\mathbf{G_2}|$. We define

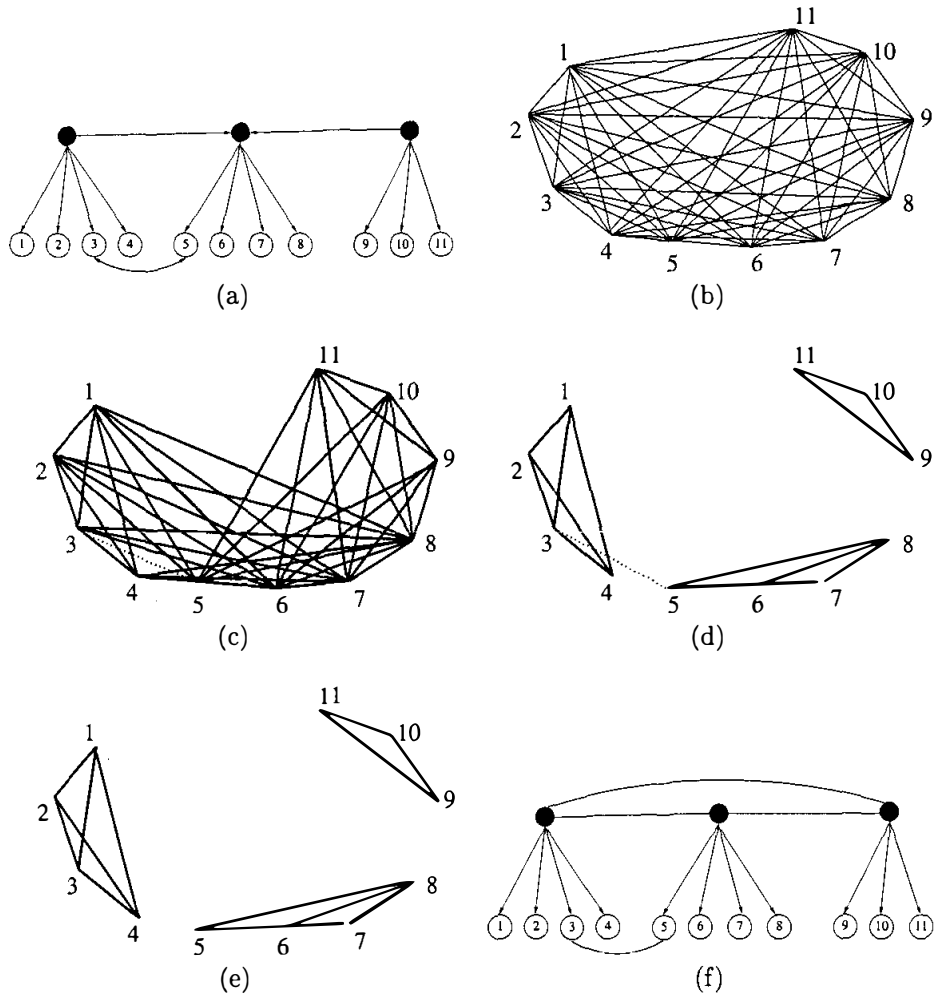

Figure 2: A step-by-step demonstration of how the graph in Figure (a) will generate the measurement pattern in Figure (f).

purifications of a measurement pattern in an analogous way purifications of latent variable graphs were defined, but only with respect to a subset of latents that should form a maximum clique within the set of latents. The following results hold with probability 1:

**Theorem 2** *Let* $G(\mathbf{L}, \mathbf{O}, \epsilon, \mathbf{E_L}, \mathbf{E_O}, \mathbf{E}_\epsilon, \mathbf{G_S})$ *be the purifiable linear latent variable graph that generates the covariance matrix* $\Sigma$ *of a set of observed random variables* $\mathbf{O}$. *Let* $MM_G$ *be the measurement pattern corresponding to the equivalence class* $MM(\mathbf{O}, \Sigma)$. *Let* $\mathbf{MM_{Pure}}$ *be the set of all purifications of* $MM_G$. *Then* $\mathbf{MM_{Pure}} =_{MM} \mathbf{G_S}$.

**Corollary 1** *For every possible pair of purifiable linear latent variable graphs* $G_1(\mathbf{L_1}, \mathbf{O}, \epsilon_1, \mathbf{E_{L_1}}, \mathbf{E_{O_1}}, \mathbf{E}_{\epsilon_1}, \mathbf{G_{S_1}})$ *and* $G_2(\mathbf{L_2}, \mathbf{O}, \epsilon_2, \mathbf{E_{L_2}}, \mathbf{E_{O_2}}, \mathbf{E}_{\epsilon_2}, \mathbf{G_{S_2}})$ *faithfully generating* $\Sigma$, *the covariance matrix of* $\mathbf{O}$, *we have* $\mathbf{G_{S_1}} =_{MM} \mathbf{G_{S_2}}$.

## 5 Complexity

The algorithms we have discussed for learning measurement patterns and pure measurement models are exponential in the worst case, since they require finding maximal and maximum cliques. The *Unclustered* test itself may require $O(n^6)$ steps, $n$ the number of variables. Such costs may limit the application of our procedure for larger problems, but in practice it will work in reasonable time if the true graph is not very impure: if the true graph contains no impurity, the procedures will run in polynomial time. In the case of the *Unclustered* test, the actual number of steps in a given problem can be much lower than $n^6$ if true clusters are relatively small with respect to the total number of variables, which can be expected as the number of nodes increases. In the same way that junction trees contributed to the development of approximate inference algorithms by providing a principled, but worst-case exponential, solution to the inference problem, the procedure outline here could be used as a starting point for creating principled approximate solutions. The sequential testing of many vanishing tetrad hypotheses may limit the confidence in the actual output. In pratice, the measurement pattern can have many errors, but still induce a correct purified solution, as our examples will illustrate.

## 6 Empirical evaluation

In Silva et al., (2003), we analyze a real-world example involving a battery of indicators of "student anxiety", and our results seem to provide further insight than those derived from different variations of factor analysis. Here we report on several simulation studies involving models with 5 to 10 latent variables and 3 to 5 indicators for each latent. We investigate true models with 1) linear relations among the latents and a pure measurement model with normal variates, 2) linear relations among the latents and an impure measurement model with normal variates, and 3) non-linear relations among the latents and an impure measurement model with non-normal variates. In studies 1 and 2 we generated the graph among the latents randomly, and samples of 1,000 and 5,000 were drawn pseudo-randomly with the Tetrad IV program[2]. Linear coefficients were uniformly sampled from the interval $[-1.5, -0.5] \cup [0.5, 1.5]$ and the variance of the exogenous nodes were uniformly sampled from the interval $[1, 3]$. The average number of neighbors for latent variables was set to 2 (in the cases of up to 5 latents) and 4 (in the case of 10 latents). The algorithm's success is evaluated by comparing the *pure* model output with respect to the maximal purified true graph (unique in the examples we generated) with the following desiderata:

- **proportion of missing latents**, the number of latents in the true graph that do not appear in the estimated pure graph, divided by the number of latents in the true graph;

- **proportion of missing measurements**, the number of indicators in the true purified graph that do not appear in the estimated pure graph, divided by the number of indicators in the true purified graph;

- **proportion of misplaced measurements**, the number of indicators in the estimated pure graph that end up in the the wrong cluster, divided by the number of indicators in the estimated pure graph;

- **proportion of impurities**, the number of impurities in the estimated pure graph divided by the number of impurities in the true (non-purified) graph. [3]

We decide which latent found by the algorithm corresponds to which of the original latents by comparing the majority of the indicators in a given estimated cluster to those in the true model: for example, suppose we have an estimated latent $L_E$. If, for instance, 70% of the measures in $L_E$ are measures of the true latent $L_2$, we label $L_E$ as $L_2$ in the estimated graph and calculate the statistics of comparison as described above. A few ties occur, but labeling the latent in one way or another did not change the final statistics.

In study 1, for a given number $m$ of latents (with random relations among them), we add $n$ pure indicators to each latent, where $m = 5, 10$ and $n = 3, 4, 5$. We used two different sample sizes: 1000 and 5000 observations. The results (Table 2), make it clear that the number of indicators contributes more to the sucess of the algorithm than the sample size. With exactly three indicators per latent, there is little margin for redundancy and any statistical mistake when evaluating a constraint may be enough to eliminate a whole cluster. There is a huge leap of quality when latents have four indicators: in this case, results are extremely good and adding more samples do not change them much. A similar pattern follows for the case with 5 and 10 latents, although the case for 10 latents, 3 indicators per latent and 5000 examples deserves further study.

In study 2, we added impure indicators to the models from study 1 prior to generating data, but the results are largely unchanged (Table 3).

The third experiment uses the graph in Figure 3 to generate data, parameterized by the following set of nonlinear structural equations among the latents:

$$\begin{aligned} L_2 &= L_1^2 + \epsilon_{L2} \\ L_3 &= \sqrt{L_1} + \epsilon_{L3} \\ L_4 &= \sin(L_2/L_3) + \epsilon_{L4} \end{aligned}$$

---

[2]Available at http://www.phil.cmu.edu/tetrad.

[3]Notice that a node that is impure in the measurement pattern may not be impure with respect to the other nodes in the purified estimated graph. In this case, we do not count them. For each pair of nodes that forms a localized impurity (e.g., indicators with correlated errors, or an indicator that is a direct cause of another, while both are children of a same and single latent), we count this pair as one impurity, since removing one of them will eliminate that impurity. Each indicator that has more than one immediate latent ancestor (i.e., a latent ancestor with a directed path to that indicator that does not include any other element in the latent set) is counted as one impurity, since it has to be removed from all purified graphs.

Table 2: Study 1. Each number is an average over 10 trials, with an indication of the standard deviation over these trials. The two columns represent the cases with 5 latents/1000 observations and 5 latents/5000 observerations, 10 latents/1000 observations and 10 latents/5000 observations, respectively.

| Evaluation of estimated purified models | | |
|---|---|---|
| | 5L/1000 | 5L/5000 |
| **3 indicators, pure** | | |
| missing latents | 0.42 ± 0.15 | 0.28 ± 0.10 |
| missing indicators | 0.36 ± 0.16 | 0.26 ± 0.10 |
| misplaced indicators | 0.11 ± 0.12 | 0.03 ± 0.08 |
| **4 indicators, pure** | | |
| missing latents | 0.0 ± 0.0 | 0.02 ± 0.06 |
| missing indicators | 0.08 ± 0.05 | 0.06 ± 0.07 |
| misplaced indicators | 0.0 ± 0.0 | 0.0 ± 0.0 |
| **5 indicators, pure** | | |
| missing latents | 0.0 ± 0.0 | 0.02 ± 0.06 |
| missing indicators | 0.03 ± 0.03 | 0.06 ± 0.08 |
| misplaced indicators | 0.0 ± 0.0 | 0.0 ± 0.0 |
| | 10L/1000 | 10L/5000 |
| **3 indicators, pure** | | |
| missing latents | 0.40 ± 0.08 | 0.45 ± 0.08 |
| missing indicators | 0.37 ± 0.09 | 0.43 ± 0.11 |
| misplaced indicators | 0.05 ± 0.08 | 0.03 ± 0.06 |
| **4 indicators, pure** | | |
| missing latents | 0.07 ± 0.08 | 0.05 ± 0.07 |
| missing indicators | 0.11 ± 0.09 | 0.10 ± 0.06 |
| misplaced indicators | 0.02 ± 0.04 | 0.0 ± 0.0 |
| **5 indicators, pure** | | |
| missing latents | 0.02 ± 0.04 | 0.0 ± 0.0 |
| missing indicators | 0.09 ± 0.07 | 0.06 ± 0.05 |
| misplaced indicators | 0.0 ± 0.0 | 0.0 ± 0.0 |

Table 3: Results for Study 2.

| Evaluation of estimated purified models | | |
|---|---|---|
| | 5L/1000E | 5L/5000E |
| **3 indicators + impurities** | | |
| missing latents | 0.40 ± 0.13 | 0.34 ± 0.16 |
| missing indicators | 0.40 ± 0.15 | 0.37 ± 0.20 |
| misplaced indicators | 0.0 ± 0.0 | 0.01 ± 0.03 |
| impurities | 0.06 ± 0.08 | 0.03 ± 0.07 |
| **4 indicators + impurities** | | |
| missing latents | 0.0 ± 0.0 | 0.04 ± 0.08 |
| missing indicators | 0.05 ± 0.08 | 0.14 ± 0.13 |
| misplaced indicators | 0.01 ± 0.01 | 0.0 ± 0.0 |
| impurities | 0.03 ± 0.09 | 0.0 ± 0.0 |
| **5 indicators + impurities** | | |
| missing latents | 0.0 ± 0.0 | 0.0 ± 0.0 |
| missing indicators | 0.05 ± 0.04 | 0.03 ± 0.03 |
| misplaced indicators | 0.0 ± 0.0 | 0.0 ± 0.0 |
| impurities | 0.03 ± 0.09 | 0.0 ± 0.0 |

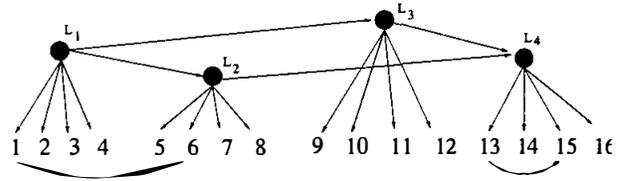

Figure 3: An impure model with a diamond-like latent structure. Notice there are two ways to purify this graph: by removing 6 and 13 or removing 6 and 15.

| Evaluation of estimated purified models | | | |
|---|---|---|---|
| | 1000 | 5000 | 50000 |
| **Wishart test** | | | |
| miss. latents | 0.20 ± 0.11 | 0.20 ± 0.11 | 0.18 ± 0.12 |
| miss. ind. | 0.21 ± 0.11 | 0.22 ± 0.08 | 0.10 ± 0.13 |
| mispl. ind. | 0.01 ± 0.02 | 0.0 ± 0.0 | 0.0 ± 0.0 |
| impurities | 0.0 ± 0.0 | 0.0 ± 0.0 | 0.1 ± 0.21 |
| **Bollen test** | | | |
| miss. latents | 0.18 ± 0.12 | 0.13 ± 0.13 | 0.10 ± 0.13 |
| miss. ind. | 0.15 ± 0.09 | 0.16 ± 0.14 | 0.14 ± 0.11 |
| mispl. ind. | 0.02 ± 0.05 | 0.0 ± 0.0 | 0.1 ± 0.03 |
| impurities | 0.15 ± 0.24 | 0.10 ± 0.21 | 0.0 ± 0.0 |

Table 4: Results obtained for the non-linear model.

where $L_1$ is distributed as a mixture of two beta distributions, $Beta(2,4)$ and $Beta(4,2)$, where each one has prior probability of 0.5. Each error term $\epsilon_{L_v}$ is distributed as a mixture of a $Beta(4,2)$ and the symmetric of a $Beta(2,4)$, where each component in the mixture has a prior probability that is uniformly distributed in $[0,1]$, and the mixture priors are drawn individually for each latent in $\{L_2, L_3, L_4\}$. The error terms for the indicators also follow a mixture of betas $(2,4)$ and $(4,2)$, each one with a mixing proportion individually chosen according to a uniform distribution in $[0,1]$. In principle, the asymptotic distribution free test of tetrad constraints from (Bollen, 1990) should be the method of choice if the data does not pass a normality test. However, such test uses the fourth moments of the empirical distribution, which can take a long time to compute if the number of variables is large (since it takes $O(mn^4)$ steps, where $m$ is the number of data points and $n$ is the number of variables). Caching a large matrix of fourth moments may require secondary memory storage, unless one is willing to pay for multiple passes through the data set every time a test is demanded or if a large amount of RAM is available. In practice, researchers may be unwilling or unable to go to the trouble. We have therefore used the Wishart test (see Spirtes et al., 2000 for details), which assumes multivariate normality. Samples of size 1000, 5000 and 50000 were used. The results (Table 4) are reasonable, and are not substantially improved by using Bollen's distribution free test.

### 6.1 Factor Analysis

For comparison, we generated factor analysis models for each of the data sets in these experiments using the PROC FACTOR procedure from SAS v.8e, and two criteria for choosing the number of latents: the default SAS criterion that chooses the number of latents by a threshold on the amount of variance explained, and an iterative procedure that chooses the number of latents by the first statistically significant model starting with 1 latent and increasing the number of latents by 1 at each iteration. Both chi-square tests and BIC scoring were used. We performed an oblique rotation (we used the oblimin rotation). We then heuristically cluster the indicators by associating each one with the latent with the respective highest loading (in absolute value). The default criterion of choosing the number of latents badly underestimated the true number. The chi-square criterion worked extremely well for the experiments with entirely linear models. The combination of the chi-square criterion and the heuristic clustering criterion achieved nearly zero error by all our evaluation measures. But in the last experiment, with a non-linear system, using samples from Figure 3, SAS worked reasonably with the default procedure, but with chi-square iteration failed to find a statistically significant model before having convergence problems with maximum likelihood estimation in 10 trials. In an actual case, we would be uncertain as to which factor analysis rotation criterion to use, and we know of no theoretical guarantees for either criterion.

## 7  Future Work

Once something can be done, it can be done in many ways. Despite a number of theoretical and practical problems, Bayesian or other score based methods could perhaps be applied, although our attempt at such an algorithm (Silva, 2002) did not perform as well. Unlike DAGs over observed variables, latent variable models cannot be decomposed (as, for instance, in Chickering (2002)), and current asymptotic approximations to the posterior distribution, such as the BIC score, are known to be inconsistent for latent variable models. One step towards solving this problem is given in Rusakov and Geiger, 2002. Factor analysis criteria could be more systematically explored, both by simulation and by theory. There are, besides, a number of possible improvements on the procedures we have described. First, we might use approximation algorithms that can handle problems with larger numbers of variables. Second, we might explore solutions for discrete variables. For instance, Bartholomew and Knott (1999) present generalizations of factor analysis to exponential family distributions, which could be used as a starting point for dealing with multinomial data under our framework. Finally, we need to do more extensive experimental evaluation, including more tests with non-Gaussian data and real-world data, as well as simulations where assumptions do not hold.